\documentclass[letterpaper, 10pt, conference]{ieeeconf}  

\IEEEoverridecommandlockouts                              

\usepackage{amsmath} 
\usepackage{amssymb} 
\usepackage{amsfonts}
\usepackage{times}  
\usepackage{balance}  
\usepackage{graphicx} 
\usepackage{blindtext}
\usepackage[hidelinks]{hyperref}
\usepackage{multirow}
\usepackage{tabularx}
\usepackage{booktabs}
\usepackage{subfig}
\usepackage{float}
\usepackage{color}
\usepackage{cite}
\usepackage[flushleft]{threeparttable}
\usepackage[linesnumbered, ruled]{algorithm2e}

\title{\LARGE \bf
Metric Monocular Localization Using Signed Distance Fields
}

\author{Huaiyang Huang$^{1}$, Yuxiang Sun$^{1}$, Haoyang Ye$^{1}$ and Ming Liu$^{1}$, \textit{Senior Member}, \textit{IEEE}
\thanks{$^{1}$The authors are with  the  Robotics  and  Multi-Perception  Laborotary, Robotics  Institute, The Hong Kong University of Science and Technology. (email: hhuangat@connect.ust.hk; eeyxsun@ust.hk, sun.yuxiang@outlook.com; hy.ye@connect.ust.hk; eelium@ust.hk).}
}

\begin{document}

\newcommand\Tstrut{\rule{0pt}{2.6ex}}         
\newcommand\Bstrut{\rule[-0.9ex]{0pt}{0pt}}   

\maketitle
\thispagestyle{empty}
\pagestyle{empty}

\begin{abstract}
    
Metric localization plays a critical role in vision-based navigation.
For overcoming the degradation of matching photometry under appearance changes, recent research resorted to introducing geometry constraints of the prior scene structure.
In this paper, we present a metric localization method for the monocular camera, using the Signed Distance Field (SDF) as a global map representation. 
Leveraging the volumetric distance information from SDFs, we aim to relax the assumption of an accurate structure from the local Bundle Adjustment (BA) in previous methods.
By tightly coupling the distance factor with temporal visual constraints, 
our system corrects the odometry drift and jointly optimizes global camera poses with the local structure.
We validate the proposed approach on both indoor and outdoor public datasets.
Compared to the state-of-the-art methods, it achieves a comparable performance with a minimal sensor configuration.

\end{abstract}


\section{Introduction}
\label{sec:intro}

Metric visual localization is a crucial building block for applications varying from surveying to autonomous navigation.
It provides location information with the absolute scale for high-level applications \cite{liu2019icmlws, wang2019self}.
With the advancement of Simultaneous Localization and Mapping (SLAM) and Structure-from-Motion (SFM) techniques \cite{zhang2014loam, ye2019tightly, yu2019omni}, localizable maps with precise structure can be built off-the-shelf, which provide fundamental prior information for global localization. 
However, how to localize a monocular camera metrically against a prior map with reliability remains an unsolved problem \cite{sun2017improving}.

Several existing methods \cite{pascoe2015direct, lynen2015get, stewart2012laps} exhibited the capability to localize camera with photometric information from LiDAR intensity, dense visual reconstruction or sparse feature maps. 
However, these approaches were subject to varying illumination and appearance changes. 
Another solution is to match geometry \cite{caselitz2016mloc, forster2013air, ykim2018sloc}. 
Under the assumption of an accurate local reconstruction from the Bundle Adjustment (BA) or stereo matching, geometry-based or hybrid approaches generally, regard visual localization as a rigid alignment problem. 
For recovering the metric scale, these methods require additional sensors that provide absolute scale information are required (e.g., depth, stereo, or inertial sensors) \cite{sun2018motion}.
Besides, the pre-built scene geometry can provide nontrivial prior  constraints for estimating the local structure, based on which we propose to perform the motion and structure optimization in a tightly-coupled manner.

\begin{figure}[t]
    \centering
    \subfloat[An example image from EuRoC dataset.]{\includegraphics[width = 0.4\textwidth]{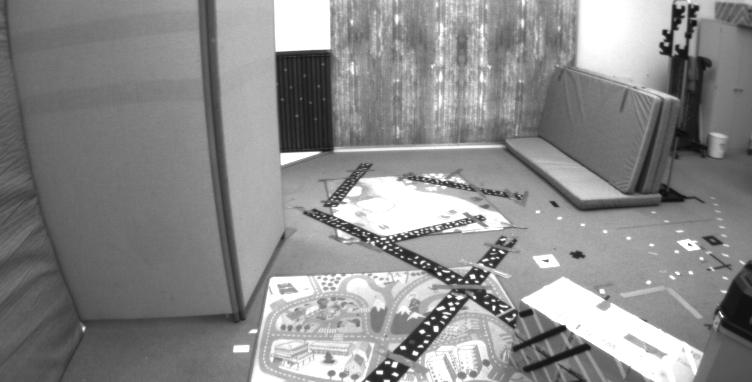}}\\
    \subfloat[Visualization of an experimental result.]{\includegraphics[width = 0.4\textwidth]{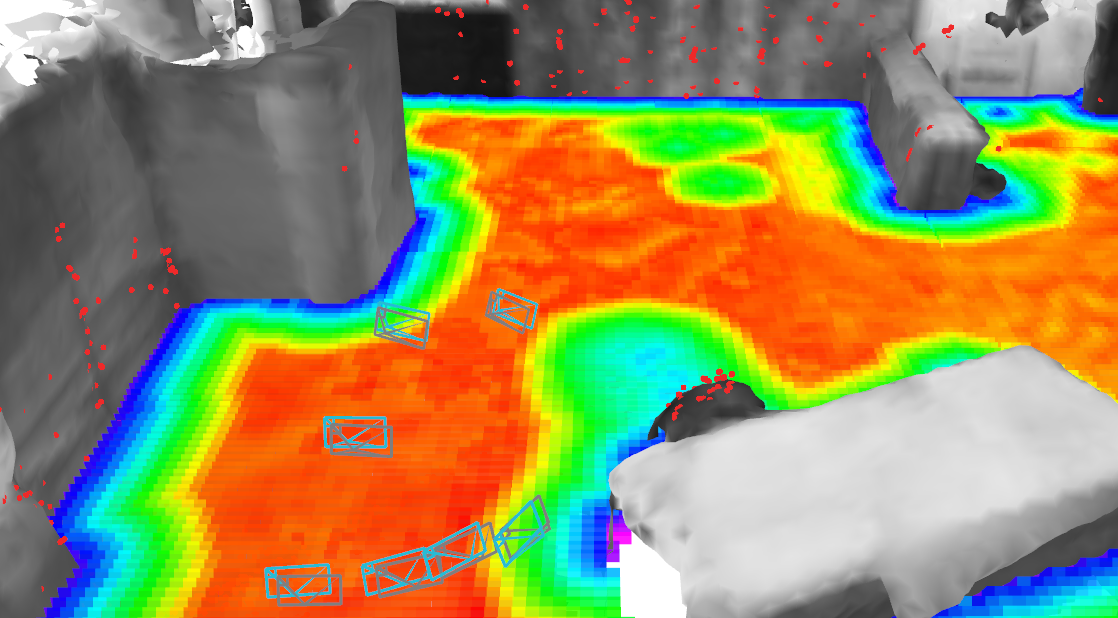}}
    \caption{An example localization result (b) in an indoor environment (a). With a prior SDF (3D voxel grid), the proposed method refines the local structure (red points) and well aligns it with the scene structure (gray mesh). The camera poses (blue frustums) are estimated accurately, compared to the groudtruth (gray frustums).}
    \label{fig:intro}
\end{figure}

Different from previous methods, we leverage the Signed Distance Field (SDF) as a global map for an accurate and robust metric localization. 
Several works \cite{klingensmith2015chisel, zucker2013chomp, bylow2013real} have proved that the SDF is a generic map representation for dense reconstruction, as well as for trajectory planning and pose estimation. 
As a SDF models an implicit scene structure, it provides direct measurements of distance along with gradients in 3D Euclidean space. 
Based on these properties, our system corrects the drift of visual odometry, retrieve depth prior for the landmark generation, and better optimize the local structure. 
\autoref{fig:intro} shows an indoor scene from EuRoC dataset \cite{Burri2016euroc} and the corresponding localization result. 
To this end, we summarize our contributions as follows:


\begin{enumerate}
    \item A hierarchical monocular localization method using the SDF as a generic map representation.
    \item A tightly-coupled approach to optimize the local structure and global poses simultaneously.
    \item Various experiments on public datasets that demonstrate the effectiveness of the proposed method and the superiority over the state-of-the-art methods.
\end{enumerate}


\section{Related Work}
\label{sec:review}

\subsection{Metric Visual Localization Against Prior Maps}
\label{sec:loc_review}

A general framework of monocular visual localization is to establish a sparse feature map with local descriptors such as SIFT \cite{lowe2004sift}. 
With the 2D-3D correspondences matching, metric poses can be estimated via a Perspective-n-Point (PnP) scheme. 
For instance, Lynen \textit{et al} \cite{lynen2015get}. proposed a lightweight visual-inertial localization system for mobile devices, which demonstrated the ability of large-scale localization. 
In contrast to an abstract representation of sparse features, a dense reconstruction with intensity makes the direct photometric matching possible. 
\cite{wolcott2014visual, stewart2012laps, pascoe2015direct} introduce Normalized Mutual Information (NMI) or its variant Normalized Information Distance (NID) to align a query image with the prior dense reconstruction with intensity.
Wolcott \textit{et al}. \cite{wolcott2014visual} proposed to search 3-DoF candidate pose hypotheses and then match their NMI with a synthetic intensity image. 
Stewart \textit{et al}. \cite{stewart2012laps} and Pascoe \textit{et al}. \cite{pascoe2015direct} utilized a closed-form formulation to avoid the exhaustive search and scoring and estimate the 6-DoF camera pose generally.
Although the NID-based similarity estimation is effective for photometric invariant localization, these methods still fail to converge in some cases due to highly non-uniform illumination.

To overcome the challenges from appearance variances, several approaches were proposed for a geometry-based or hybrid solution. 
In \cite{forster2013air}, Foster \textit{et al}. presented a Monte Carlo Localization (MCL) method from correlating the height-maps originated from the alignment of the local structure with depth information. 
Similarly, Caselitz \textit{et al}. \cite{caselitz2016mloc} proposed to match the local reconstruction from visual odometry with the prior pointcloud through estimating a similarity transform. 
Kim \textit{et al}. \cite{ykim2018sloc} proposed to directly minimize the depth residual between the stereo disparity map and depth map projected from 3D LiDAR information. 
For these methods, stereo or depth sensors are required to acquire depth information \cite{forster2013air,ykim2018sloc}, or manually initialized the scale from the groundtruth \cite{caselitz2016mloc}. 
Furthermore, while a prior map is supposed to contribute to the structure optimization in local BA, current works are generally based on the assumption that the local structure is accurate. 
We argue that it is beneficial to introduce such factors to optimization procedures, including both the local structure and global camera poses, for a robust and drift-free visual localization.

\subsection{SDFs for Visual Navigation}
\label{sec:review_sdf}

SDF was firstly introduced in computer vision as a map representation for dense reconstructions \cite{gibson1998using}. 
For instance, KinectFusion \cite{newcombe2011kinectfusion} is an accurate RGB-D reconstruction framework modelling the surroundings with volumetric Truncated Signed Distance Field (TSDF). 
Voxblox \cite{oleynikova2017voxblox} managed to build Euclidean Signed Distance Fields (ESDF) with heterogeneous sensor types for Micro Aerial Vehicles (MAVs).
As SDFs provide metric information of static obstacles, they have long been used for planning. Chomp \cite{zucker2013chomp} was one of the state-of-the-art gradient-based planning algorithms using SDFs. 

Besides, several methods attempted to exploit SDFs for camera pose estimations \cite{forster2013air, klingensmith2015chisel, bylow2013real}. 
They either aligned the structure with SDF, or resorted to SDF for local reconstructions. 
Bylow \textit{et al}. \cite{bylow2013real} proposed to a RGB-D Simultaneous Localization and Mapping (SLAM) framework. 
This method directly tracked camera poses by minimizing a cost function associated with the current TSDF reconstruction. Foster \textit{et al}. 
\cite{forster2013air} used the SDF as a penalty function for a better multi-view reconstruction, which served as the first step for camera localization. 
Existing works generally assumed an accurate structure and then tracked the camera pose by rigid structure alignment. 
On the contrary, we consider a tightly-coupled monocular localization problem with the SDF. 



\begin{figure}[t!]
	\centering
	\includegraphics[width=0.45\textwidth]{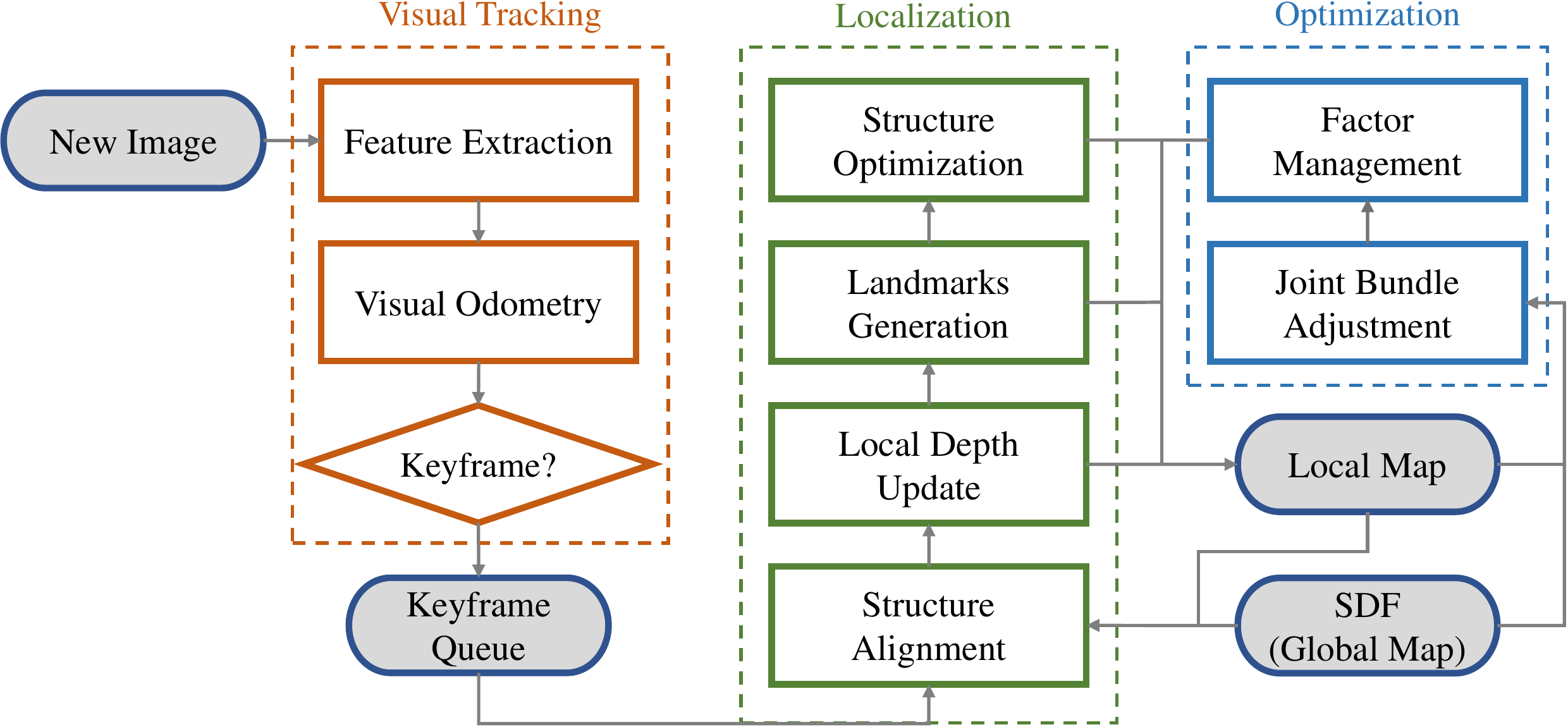}
	\caption{Framework of the proposed system.}
	\label{fig:frame}
\end{figure}


\section{Overview}
\label{sec:overview}

\autoref{fig:frame} shows the diagram of the proposed system. 
The proposed localization system is implemented with a \textit{track-then-localize} scheme.
Firstly, the pose of the current camera is tracked by the visual odometry front-end.
With an initial local structure, we use a visual odometry framework to accomplish frame-to-frame and frame-to-model tracking. 
Motion and structure BA is then performed with SDF constraints to localize the camera and align the scene geometry.
In the meantime, the back-end jointly optimizes the global camera poses along with the local structure. 

The key idea of our method is to optimize the local structure and global poses simultaneously, utilizing the volumetric SDF as an global constraint. 
The SDF, denoted as $\boldsymbol{\phi} : \mathbb{R}^3 \rightarrow \mathbb{R}$, implies an isosurface $\boldsymbol{\phi}(\mathbf{\mathbf{p}}) = 0$ representing the pre-built world model. 
For any point $\mathbf{p}$ in the 3D Euclidean space, SDF provides a distance measurement $\boldsymbol{\phi}(\mathbf{p})$ to the nearest isosurface, along with a smooth gradient $\nabla \boldsymbol{\phi}(\mathbf{p})$. 
Based on these properties, the structure could be aligned and refined with this implicit model for global localization.

Throughout this paper, we denote camera poses as matrices of the special Euclidean group $\mathbf{T}_k\in \mathbf{SE}(3)$, which transform a 3D point $\mathbf{p}_i$ from the world coordinate to the camera coordinate by $\mathbf{p}^c_i = \mathbf{T}_k \cdot \mathbf{p}_i$. The update of a camera pose with an incremental twist $\boldsymbol{\xi} \in \mathfrak{se}(3)$ using a left-multiplicative formulation $\oplus : \mathfrak{se}(3)\times \mathbf{SE}(3)\rightarrow \mathbf{SE}(3)$, denoted as:
\begin{equation}
	\boldsymbol{\xi}\oplus\mathbf{T}_{k}:=\exp ( \boldsymbol{\xi}^\wedge )\cdot \mathbf{T}_k .
\end{equation}

The corresponding pixel of a 3D point $\mathbf{p}_i$ observed in the $k$-th image, is denoted as $\mathbf{u}_{i, k} = (u, v)^T \in \mathcal{I}$ . 
And we use $\pi:\mathbb{R}^3 \rightarrow \mathbb{R}^2$ to denote the camera projection model, which transform a point to the image coordinate by: 
\begin{equation}
	\mathbf{u_{i, k} = \pi(\mathbf{p}_i^c)} .
\end{equation}

\section{Methodology}
\label{sec:method}

\subsection{Visual Tracking}

To demonstrate the capability of tightly coupling with existing visual odometry algorithms, here we adopt built-in modules of ORB SLAM2 \cite{mur2017orb} for a tracking front-end. 
The tracking module extracts FAST corners and computes their ORB descriptors \cite{rublee2011orb}, which are tracked with the previous frame via matching the descriptors in a local window. 
The camera pose is then estimated via a PnP scheme. 
After the visual tracking, all the essential information is delivered to the following modules for localization, including the features extracted, the pose of the current frame and 3D correspondences that are tracked successfully.

We separate the local structure into two sets according to their observation sources. 
We denote $\mathcal{P} = \left\{\mathbf{p}_i\right\}$ with $\mathbf{p}_i\in\mathbb{R}^3$ as the set of all 3D features belonging to the local structure. 
We divide $\mathcal{P}$ into two separate sets as: $\mathcal{P} = \mathcal{M}\cup\mathcal{N}$, 
where $\mathcal{M}$ represents for points has a constraint with the pre-built map and $\mathcal{N}$ represents for points only constrained by multi-view stereo. 

\begin{figure}[t!]
	\centering
	\includegraphics[width=0.45\textwidth]{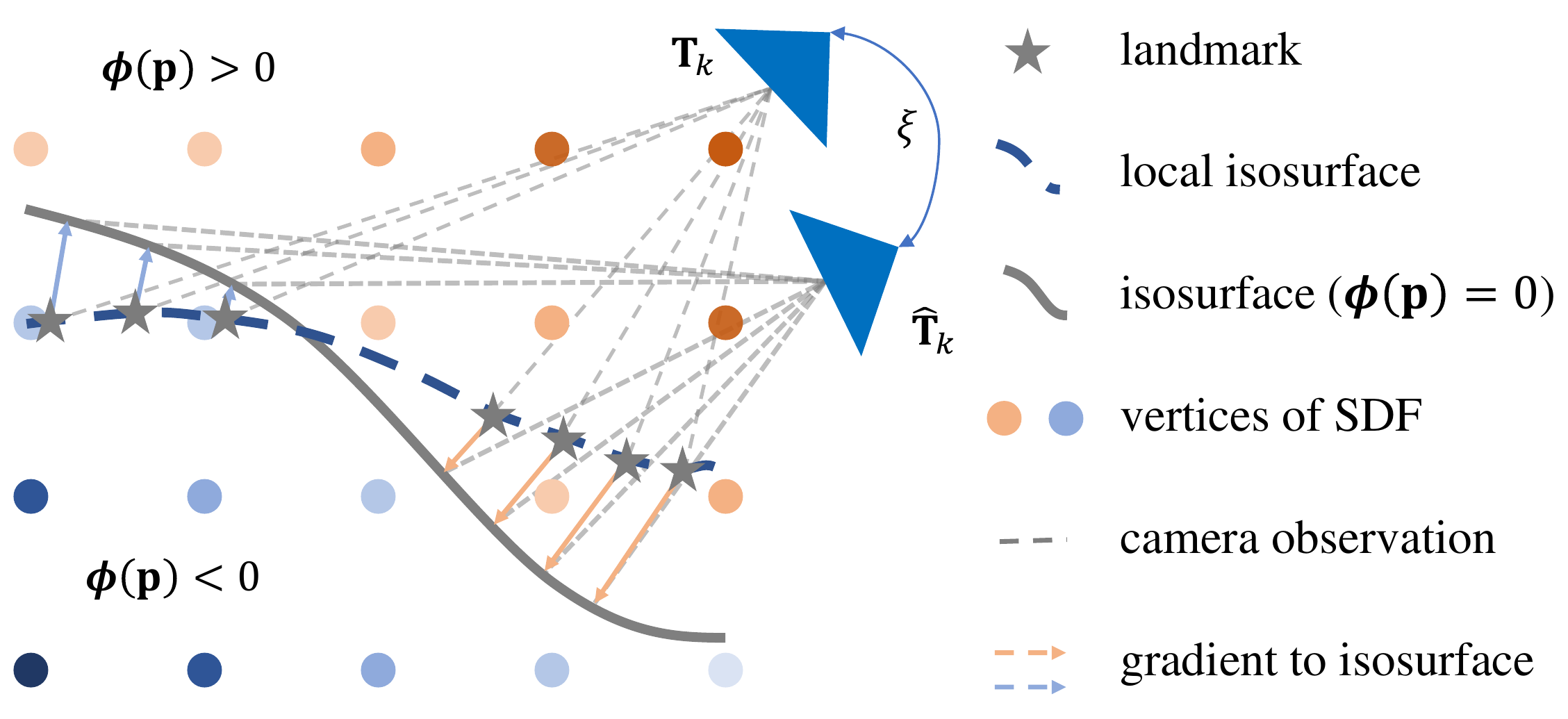}
	\caption{Illustration for pose refinement. The implicit structure is inferred from local sparse features. With the smooth gradient at each landmark pointing to global isosurface, we align the local structure with map and the camera pose is corrected seamlessly.}
	\label{fig:align}
\end{figure}

\subsection{Pose Refinement}
\label{sec:pose_refine}

To eliminate the inevitable drift of visual odometry over time, we refine the pose of current keyframe with map constraints. 
\autoref{fig:align} illustrates this localization process. 
Similar to \cite{bylow2013real}, for every $\mathbf{p}_i\in\mathcal{M}_k$, we define an energy function measures the fitness of local structure with the map, which is given by:
\begin{equation}
    e_i^{sdf} =  \boldsymbol{\phi} \left( \mathbf{p}^\prime\right) = \boldsymbol{\phi} \left( \left(\boldsymbol{\xi}\oplus\mathbf{T}_{k}\right) \mathbf{p}_i\right) , 
\end{equation}
where $\mathbf{p}^\prime$ is the position of $\mathbf{p}$ in camera coordinate. 
$\mathcal{M}_k \subset \mathcal{M}$ is the features tracked in the current frame with SDF constraints. 
$\mathbf{T}_k \in \mathbf{SE}(3)$ is initially estimated by the tracking module. 
As the optimal solution is generally not far from the initial guess, we use an incremental update policy $\hat{\mathbf{T}}_k \leftarrow \boldsymbol{\xi}\oplus\mathbf{T}_k$ to correct the trivial drift. 
The update is parametrized as a twist $\boldsymbol{\xi} \in \mathfrak{se}(3)$ and is set to be zero at the start of the pose optimization. 

The optimal pose is solved via minimizing the full energy function over all the points constrained by the SDF, formulated as:
\begin{equation}
    \hat{\boldsymbol{\xi}} = \arg \min_{\boldsymbol{\xi}}\sum_{\mathbf{p}_i\in\mathcal{M}_k}\|  \boldsymbol{\phi} \left( \left(\boldsymbol{\xi}\oplus\mathbf{T}_{k}\right) \mathbf{p}_i\right) \|_{\gamma}, 
\end{equation}
where $\|\|_{\gamma}$ denotes robust Huber norm for convergence regardless of outliers. And the twist is updated by:
\begin{equation}
    \boldsymbol{\boldsymbol{\xi}} \leftarrow \log \left(\delta \boldsymbol{\xi} \oplus \mathbf{T}(\boldsymbol{\xi})\right).
\end{equation}





We use Levenberg-Marquardt algorithm to solve $\delta\boldsymbol{\xi}$ iteratively. The system is represented as:
\begin{equation}
	\label{eq:delta}
	\delta\boldsymbol{\xi} = \mathbf{H}^{-1}\mathbf{b},
\end{equation}
\begin{equation}
	\label{eq:hessian}
    \mathbf{H} = \mathbf{J}^T\boldsymbol{\Omega}^{-1}\mathbf{J}+\beta \mathbf{I} \quad\text{and} \quad \mathbf{b}=-\mathbf{J}^T\boldsymbol{\Omega}^{-1}\mathbf{r},
\end{equation}
where the $\boldsymbol{\Omega}$ is the covariance matrix calculated from the pre-defined weight, described in \autoref{sec:new_pts}. And 
\begin{equation}
	\label{eq:ja}
    \begin{split}
        \mathbf{J}_i & = \mathbf{J}_{\text{sdf}} \cdot \mathbf{J}_{\text{pose}} \\
        & = \frac{\partial \boldsymbol{\phi} \left(\mathbf{p}^\prime \right)}{\partial\mathbf{p}^\prime}\cdot\frac{\partial \left(\left(\boldsymbol{\xi} \oplus \mathbf{T}_k \right)\mathbf{p}_i\right)}{\partial\boldsymbol{\xi}}, \\
    \end{split}
\end{equation}
where $\mathbf{J}_{\text{sdf}} = \nabla \boldsymbol{\phi}\left(\mathbf{p}^\prime\right)$ is the three-dimensional gradient indicating the direction of the nearest isosurface at global position $\mathbf{p}^\prime$. We use trilinear interpolation between neighbourhood vertices to calculated an approximate gradient for each $\mathbf{p}^\prime$ at every optimization step. 



\subsection{Landmark Generation and Depth Update}
\label{sec:new_pts}

\begin{figure}[t!]
	\centering

	\includegraphics[width=0.48\textwidth]{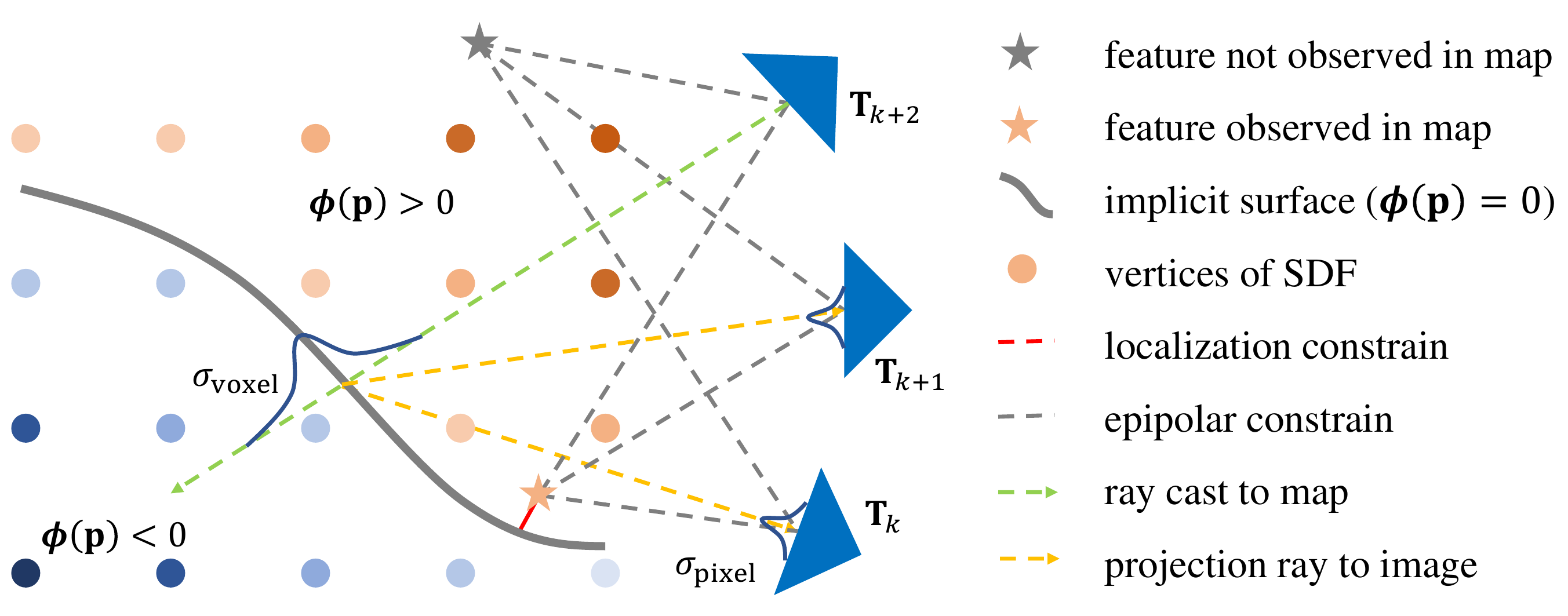}

	\caption{Illustration for landmark generation and depth update. If we find zero-crossing voxel (green ray), the feature is matched via epipolar-line search (orange ray) and the depth will be initialized according to the variances (blue curves). If not, we triangulate the 3D position (gray feature) by temporal matching only (gray rays). The structure is then optimized combining visual (gray dashed line) and SDF factors (red line).}
	\label{fig:new_pt}
\end{figure}
New landmarks are generated in two steps. The first step is to lookup the zero-crossing voxel along the optical ray. For an efficient query, we define a visual frustum according to camera intrinsics. 
The voxels are stored in a hash-table to support fast loopup \cite{oleynikova2017voxblox}.

As shown in \autoref{fig:new_pt}, if the ray is cast successfully, the depth $d_{\mathbf{p}}$ of current feature is computed by trilinear interpolation and is assumed with a variance of one pixel. 
By searching along the epipolar line, temporal matching is then performed with all the keyframes that have overlap in camera view. 
If ray-casting fails, we directly perform temporal feature matching across frames and compute landmark positions via mid-point triangulation. Landmarks will be inserted to $\mathcal{M}$ if it could be observed from the SDF and $\mathcal{N}$ if not.


On the contrary to methods based on rigid registration, the local structure will be refined to better fit the epipolar constraints and the implicit global structure simultaneously, through minimizing the energy function in \autoref{eq:energy}. To balance the relative weights of temporal multi-view stereo and the SDF, a coupling factor $\lambda$ is introduced, thus the energy function is given as:
\begin{equation}
	\label{eq:energy}
    E = E_{\text{repro}} + \lambda \cdot E_{\text{sdf}},
\end{equation}
where 
\begin{equation}
    E_{\text{sdf}} = \sum_{\mathbf{p}_i\in\mathcal{M}} e_i^{\text{sdf}}(\mathbf{p}_i) = \sum_{\mathbf{p}_i\in\mathcal{M}} w_i^{\text{sdf}} \| \boldsymbol{\phi}(\mathbf{p}_i) \|_\gamma
\end{equation}
is the energy function over the distance penalty of all the landmarks observed in the map and 
\begin{equation}
	\begin{split}
	E_{\text{repro}} &= \sum_{\mathbf{p}_i \in \mathcal{P}} \sum_{j \in \mathcal{F}_i} e_{ij}^{\text{repro}}(\mathbf{p}_i, \mathbf{T}_j) \\
	 &= \sum_{\mathbf{p}_i \in \mathcal{P}} \sum_{j \in \mathcal{F}_i} w_{i,j}^{\text{repro}}\| \textbf{u}_{i,j} - \pi(\mathbf{T}_j\mathbf{p}_i)\|_{\gamma} \\
	\end{split}
\end{equation}
defines the visual constraints generated by temporal multi-view stereo. $\mathcal{F}_i$ is the set of keyframes in which $\mathbf{p}_i$ could be observed. 
$w_{i,j}^{\text{repro}}$ is dependent on the location variance of feature on the image plane, which is set to be one-pixel at each level of the image pyramid. 
$w_i^{sdf}$ is associated with the uncertainty of the prior volumetric map and therefore we define it as $ w_{\text{sdf}}= 1/{\sigma_{\text{sdf}}^2}$. 
This weighting in the meantime normalize the reprojection error and distance error, and thus $\lambda$ becomes the only factor that makes a balance between both factors.
We perform an optimization procedure similar to \autoref{sec:pose_refine} to achieve the optimal local structure $\hat{\mathcal{P}}$.


\subsection{Joint Optimization}
\label{sec:opt}



For an accurate and drift-free localization, we use a stand-alone backend to jointly optimize the global poses and the local structure. The overall energy function is the same as \autoref{eq:energy} with a different set of optimization parameters, denoted as $\mathcal{X} = (\mathcal{P}, \mathcal{T})$, where $\mathcal{T}$ consists of the global poses for all the keyframes in the current optimization window. The incremental update is computed similar to \autoref{eq:delta} and \autoref{eq:hessian}. For simplification we assume the $k$-th residual term is relevant to $i$-th feature point $\mathbf{p}_i$ and $j$-th keyframe $\mathbf{T}_j$, thus the associated row $\mathbf{J}_k$ of Jacobian $\mathbf{J}$ defined as:  
\begin{equation}
	\mathbf{J}_k = [\lambda\mathbf{J}_{\text{sdf}} + \mathbf{J}_{\text{repro}}, \mathbf{J}_{\text{pose}}],
\end{equation}
where $\mathbf{J}_{\text{sdf}}$ is the same in \autoref{eq:ja} and $\mathbf{J}_{\text{repro}} = \partial e_{ij}^{\text{repro}}/{\partial\mathbf{p}}$, $\mathbf{J}_{\text{pose}}=\partial e_{ij}^{\text{repro}} /\partial \mathbf{T}_j$ respectively. The optimal parameters are represented as:
\begin{equation}
	\hat{\mathcal{X}} = \arg \min_{\mathcal{X}}E_{\text{repro}} + \lambda \cdot E_{\text{sdf}}.
\end{equation}

A two-step optimization is performed for better filter out the outliers. In the first round, the system will converge regardless of outliers. Then outliers are detected and rejected according to Chi-Squared test. Ignoring the outliers, we perform another round of optimization. Finally, the outliers with large residuals will be determined and dismissed.


Due to the artifacts in pre-built map, non-converged points might be an inlier but is not able to fit the global structure. Therefore, we keep a tolerant occlusion strategy for outliers. If $ w_i^{\text{sdf}}\|e_{i}^{\text{sdf}}(\mathbf{p}_i)\|_\gamma > th_{\text{sdf}}$, instead of classifying $\mathbf{p}_i$ as an outlier, we remove the factor constrained by SDF in the optimization. If in the next optimization step, the related visual factor $ w_{i, j}^{\text{repro}}\|e_{i, j}^{\text{repro}}(\mathbf{p}_i, \mathbf{T}_{j})\|_\gamma> th_{\text{\text{repro}}}$ for any $j \in \mathcal{F}_i$, $\mathbf{p}_i$ is considered as an outlier and all the relevant observation and factors will be removed from the system. Similar to \cite{mur2017orb}, we set the outlier rejection thresholds via $\mathcal{X}^2$-test, with the assumption of 95\% inliers and 1-pixel or 1-volume variance for visual factor and SDF factor respectively. Accordingly, we set $th_{\text{sdf}} = 3.841$ and $th_{\text{repro}} = 5.991$.
















\section{Experimental Results and Discussions}
\label{sec:exp}

We evaluated our algorithm on public datasets EuRoC \cite{Burri2016euroc} and KITTI Odometry \cite{Geiger2012kitti}. We used Voxblox \cite{oleynikova2017voxblox} for the volumetric SDF reconstruction. For EuRoC dataset, the global maps were established from dense stereo depth map and ground truth poses, with the voxel size $\Delta_{\text{voxel}}=0.1m$. For KITTI odometry, along with ground truth poses, Velodyne HDL-64E LiDAR scans within a radius of $25m$ were extracted for SDF reconstruction. 
As we did not leverage complex dense stereo reconstruction techniques for a consistent and high-accuracy map, some artifacts were manually corrected. 
For all the sequences, variance of SDF $\sigma_{\text{sdf}}$ was set to be identical as $\Delta_{\text{voxel}}$. The coupling factor $\lambda$ was tuned manually dependent on the scene depth and camera fps. For indoor sequences of EuRoC, we set $\lambda=1.0$. For a large-scale outdoor environment with relatively low camera frequency such as KITTI, $\lambda$ was set to $10.0$ to emphasize localization constraints. 

We first compared the general localization performance against the state-of-the-art methods on both datasets to validate the effectiveness of the proposed method. 
To further demonstrate the capability of refining the local structure and initialization with a coarse initial guess, we then evaluated the accuracy of the sparse feature map and the consistency in localization performance against different initial guesses.




\subsection{Localization Results}
\label{sec:loc_acc}


\begin{table}[htbp]
	\caption{Comparison of Translation ATE (m) on EuRoC. The proposed method is compared against state-of-the-art localizers (marked as *). }
	\begin{center}
		\begin{tabular}{cccc}
			\hline
			           & \textbf{MH01} & \textbf{V201} & \textbf{V101}                                                          \Tstrut \\
			           & (Map: MH02)   & (Map: V202)   & (Map: V102)      \Bstrut                                                       \\
			\hline
			Ours*  & 0.032         & 0.041         & 0.034    \Tstrut                                                               \\
			ROVIOLI*   & 0.082         & 0.057         & 0.110                                                                          \\
			ORB Loc*   & 0.464         & X             & X                  \Bstrut                                                     \\

			\hline
			VINS-Mono  & 0.12         & 0.081        & 0.068                    \Tstrut                                               \\
			ORB SLAM2 & 0.037        & 0.035        & 0.035  \Bstrut                                                                 \\
			\hline
		\end{tabular}
		\label{tab:acc_euroc}
	\end{center}
\end{table}

\subsubsection{The EuRoc MAV Dataset}
The EuRoC MAV Dataset provides sequences of stereo images and IMU data streams in three different indoor scenes, with extrinsic information, ground truth poses and structures. 
To compare our algorithm against the state-of-the-art methods, for each stage we built a localizable map on one sequence, and then evaluated the localization accuracy on the other. \autoref{tab:acc_euroc} shows the localization results on the three scenes.

We compare the Root Mean Square Error (RMSE) of Absolute Trajectory Error (ATE) for the translation against ROVIOLI \cite{schneider2018maplab} and ORB SLAM2 in localization mode (denoted as ORB Loc). 
Due to illumination and viewpoint variances, ORB Loc fails on two of the sequences (marked as X).
Along with the comparison of global localization, we list results of state-of-the-art visula-inertial, VINS-Mono \cite{qin2018vins}, and stereo SLAM , ORB\_SLAM2 \cite{mur2017orb}, to prove that our method improves the accuracy of local monocular odometry.

\begin{figure}[t!]
	\centering

	\subfloat{\includegraphics[width=0.208\textwidth]{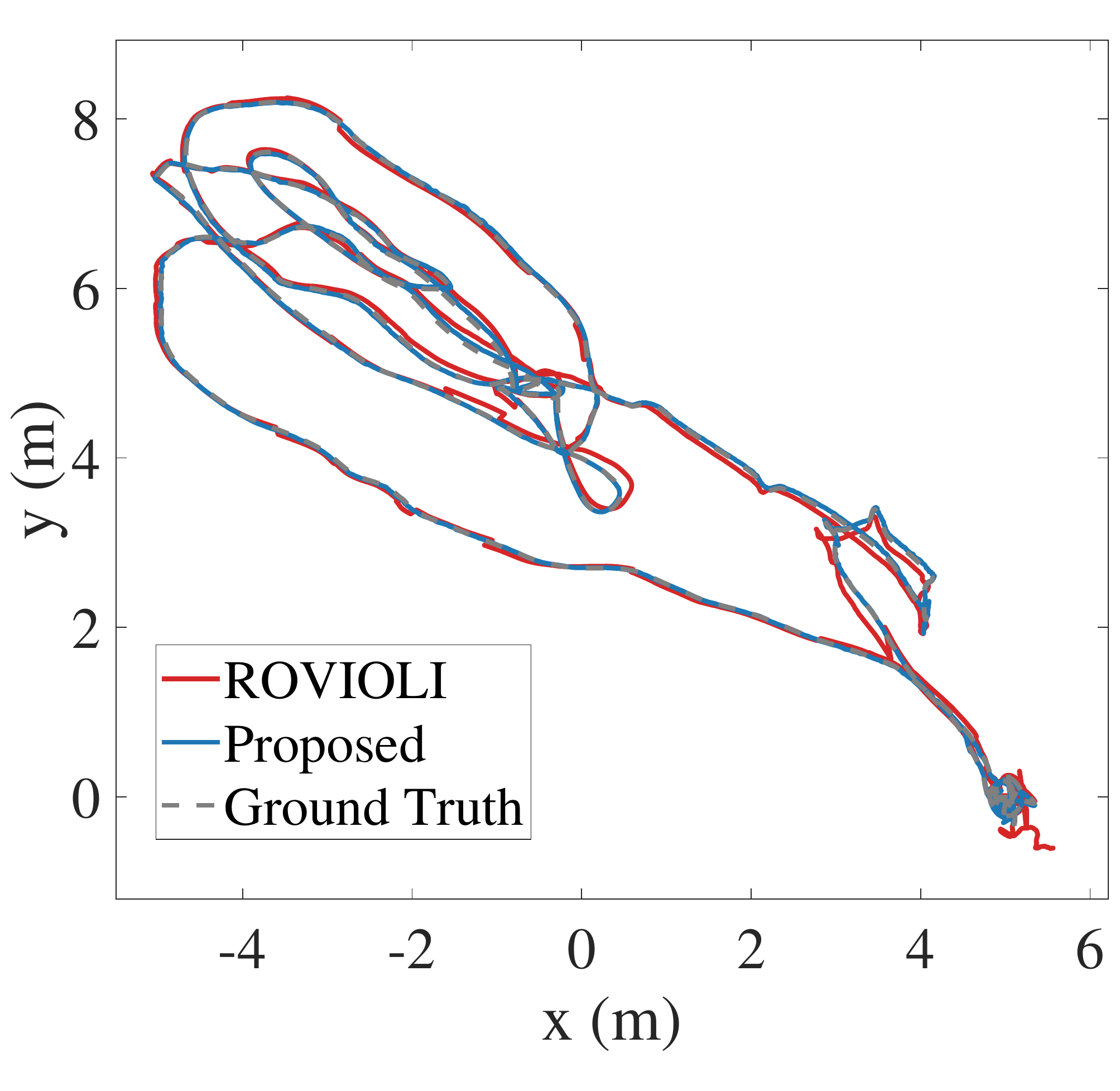}}
	\quad \subfloat{\includegraphics[width=0.21\textwidth]{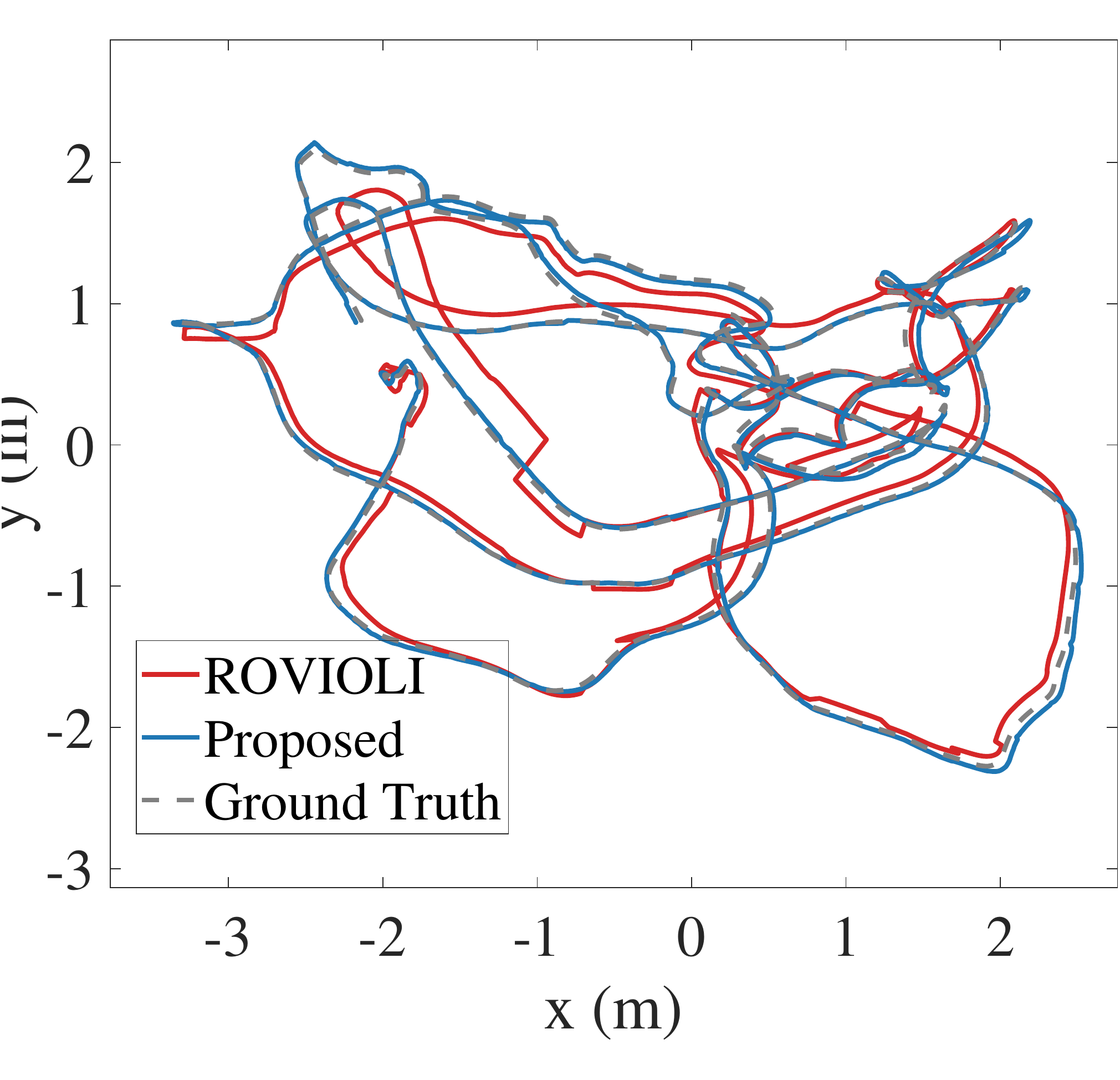}}

	\caption{Trajectory comparison on EuRoC. Left: MH01; Right: V101.}
	\label{fig:traj_euroc}
\end{figure}


 For all the three sequences, the localization error of the proposed method is better than ROVIOLI, indicating a more accurate global pose estimation, with an average RMSE lower than 0.04m. In addition, our method achieves comparable pose estimation results against VINS-Mono and ORB SLAM2. This comparison indicates that our approach can automatically recover the scale powered by combining the SDF cost with the local reprojection error.

\begin{table*}[ht]
	\caption{Comparison of ATE on KITTI. The ATE of each sequence is shown as $mean\pm std$.}
	\begin{center}
		\begin{tabular}{cccccccc}
			\hline
			\multirow{2}{*}{\textbf{Seq \#}} & \multirow{2}{*}{\textbf{Environments}} & \multicolumn{2}{c|}{\textbf{Ours}} & \multicolumn{2}{c|}{\textbf{Stereo Localizer} \cite{ykim2018sloc}} & \multicolumn{2}{c}{\textbf{ORB SLAM2 (Stereo)}} \Tstrut                                                                                                                                                  \Bstrut  \\
			\cline{3-8}
			                                 &                                        & \textbf{\textit{Translation} (m)}         &\multicolumn{1}{c|}{ \textbf{\textit{Rotation} (deg)}}                    & \textbf{\textit{Translation} (m)}                           &\multicolumn{1}{c|}{ \textbf{\textit{Rotation} (deg)}}        & \textbf{\textit{Translation} (m)}                                           & \textbf{\textit{Rotation} (deg)} \Tstrut\Bstrut \\
			\hline
			01                               & Highway                                & \textbf{5.66} $\pm$ \textbf{3.24}     & \multicolumn{1}{c|}{1.55 $\pm$ \textbf{0.49}}                      & \multicolumn{2}{c|}{X}                                   & 10.54 $\pm $ 3.37                 & \textbf{1.44} $\pm$ 0.59                                        \Tstrut                                      \\
			03                               & Country                                & 1.59 $\pm$ 0.92                       & \multicolumn{1}{c|}{0.97 $\pm$ \textbf{0.16}}                      & \textbf{0.24} $\pm $ \textbf{0.23}                      & \multicolumn{1}{c|}{\textbf{0.41} $\pm$ 0.33} & 0.59 $\pm$ 0.35                                                         & 0.60 $\pm$ 0.17                    \\
			04                               & Country                                & 0.42 $\pm$ 0.27                       & \multicolumn{1}{c|}{\textbf{0.81} $\pm$ \textbf{0.07}}             & 0.45 $\pm$ 0.35                                         & \multicolumn{1}{c|}{0.88 $\pm$ 0.39}                   & \textbf{0.17} $\pm$ \textbf{0.07}                                       & 0.99 $\pm$ \textbf{0.07}           \\
			06                               & Urban                                  & \textbf{0.22} $\pm$ \textbf{0.12}     & \multicolumn{1}{c|}{\textbf{0.32} $\pm$ 0.39}                      & 0.38 $\pm$ 0.67                                         & \multicolumn{1}{c|}{0.85 $\pm$ 1.80}                   & 0.55 $\pm$ 0.19                                                         & 0.40 $\pm$ \textbf{0.09}  \\
			07                               & Urban                                  & 0.28 $\pm$ 0.26                       & \multicolumn{1}{c|}{\textbf{0.18} $\pm$ \textbf{0.08}}             & \textbf{0.13} $\pm$ \textbf{0.12}                       & \multicolumn{1}{c|}{0.49 $\pm$ 0.47}                   & 0.40 $\pm$ 0.15                                                         & 0.39 $\pm$ 0.18                    \\
			10                               & Urban+Country                          & 0.34 $\pm$ \textbf{0.26}              & \multicolumn{1}{c|}{\textbf{0.36} $\pm $ \textbf{0.25}}            & \textbf{0.24} $\pm$ 0.68                                & \multicolumn{1}{c|}{0.49 $\pm$ 0.77}                   & 0.90 $\pm$ 0.40                                                         & 0.90 $\pm$ 0.36        \Bstrut     \\
			\hline
		\end{tabular}
		\label{tab:eval_kitti}
	\end{center}
\end{table*}


\subsubsection{The KITTI Odometry Dataset}

We validated our algorithm on the KITTI odometry dataset. For sequences containing loopy trajectories, the groundtruth poses lack such loop-closure consistency. Therefore, we only estimated the results on the loop-less sequences. We compared the average translation and rotation error along with the variances against a recently published stereo localizer \cite{ykim2018sloc} and stereo SLAM \cite{mur2017orb}.

\begin{figure}[t!]
	\centering
	\subfloat[]{\includegraphics[width=0.42\textwidth]{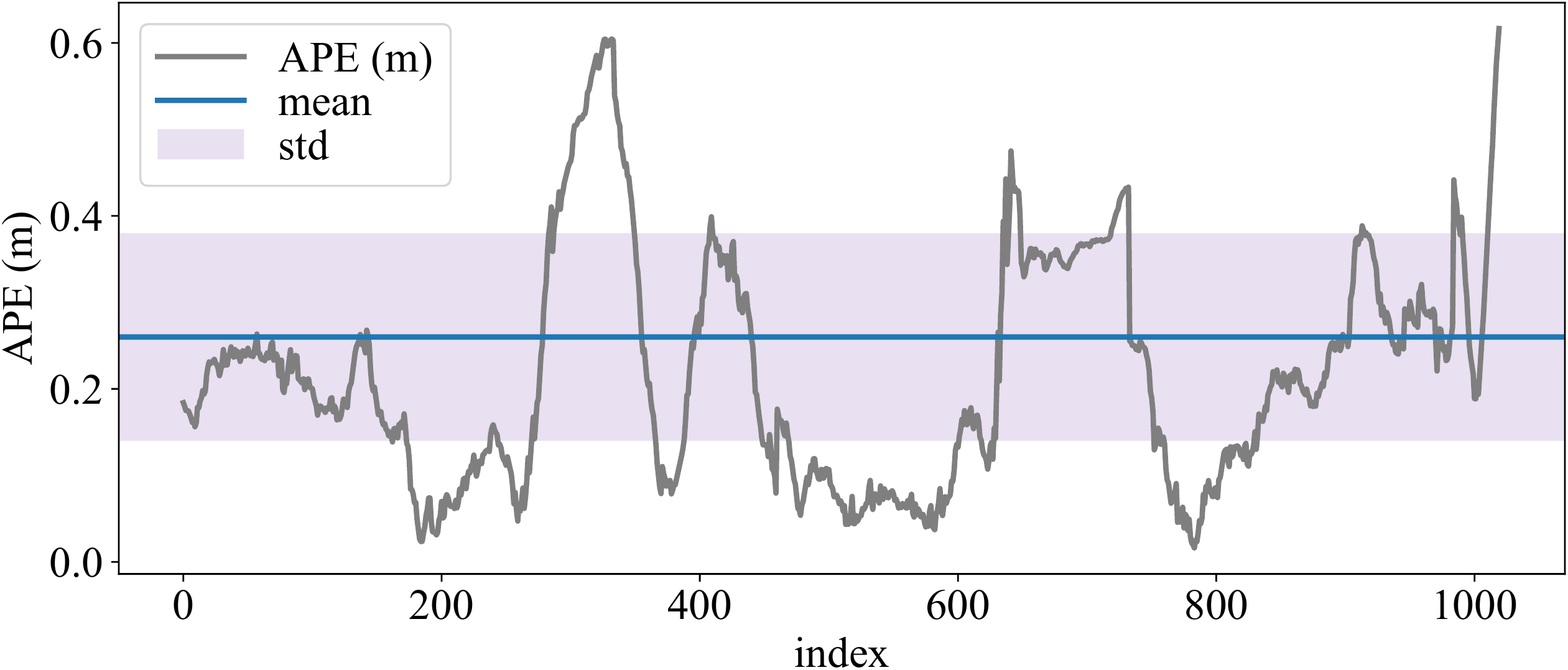}} \\
	\subfloat[]{\includegraphics[width=0.42\textwidth]{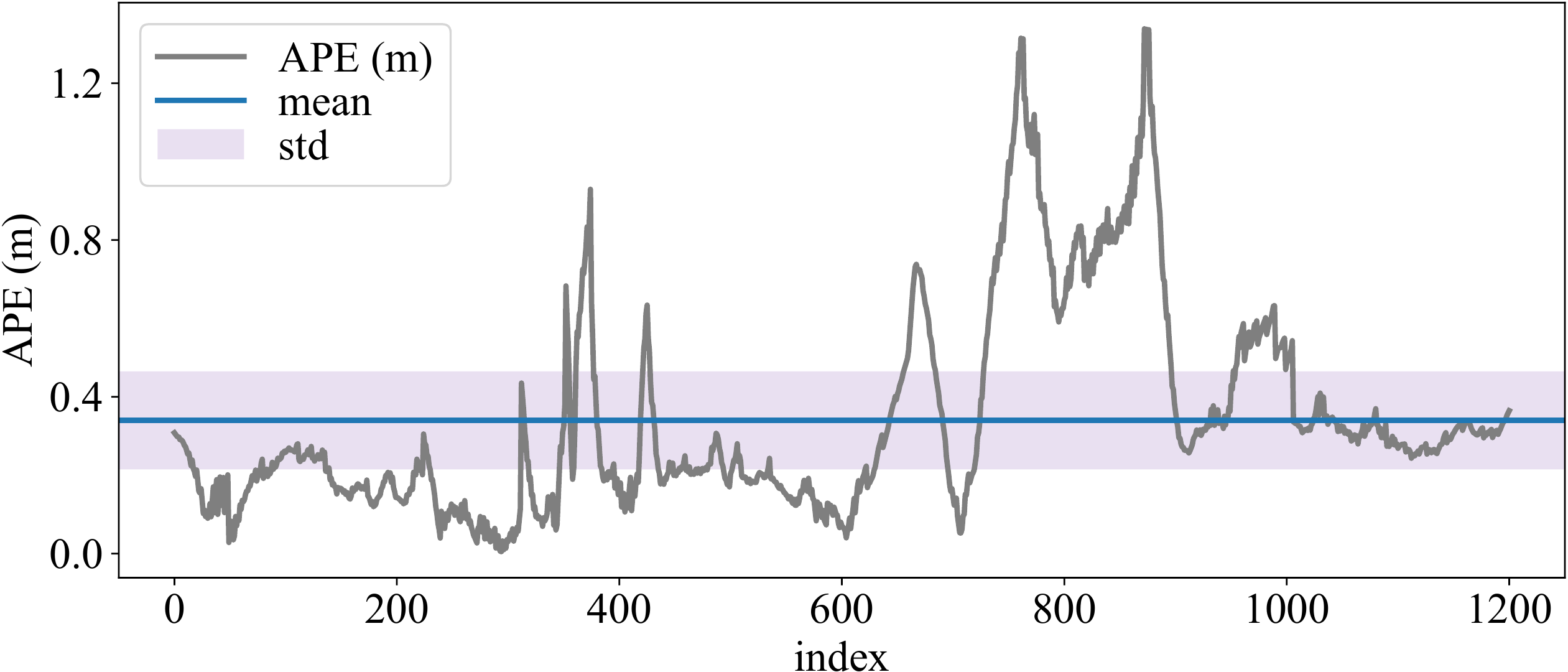}} \\
	\subfloat[]{\includegraphics[width=0.22\textwidth]{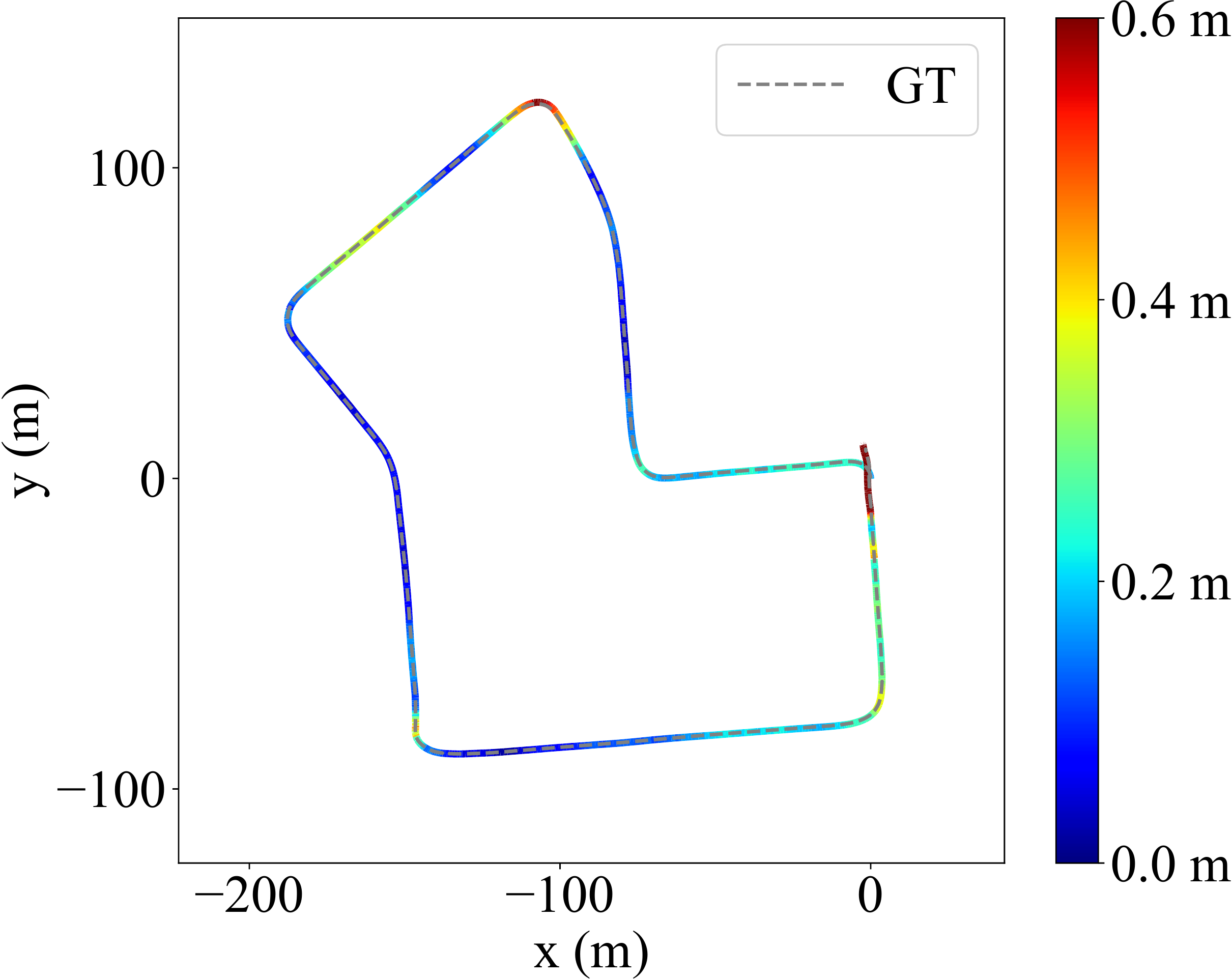}} \enspace
	\subfloat[]{\includegraphics[width=0.22\textwidth]{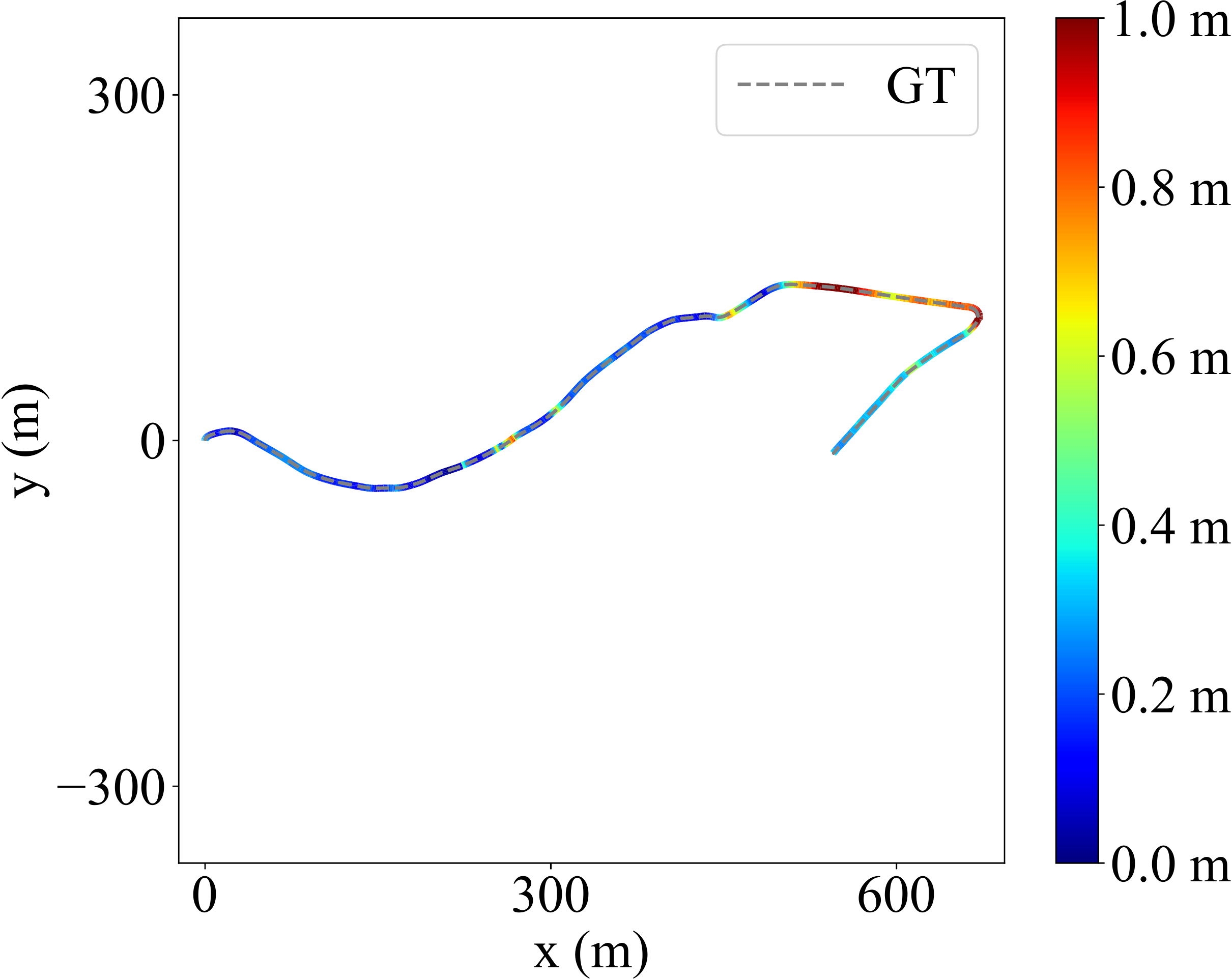}}
	\caption{Localization error on KITTI. (a) and (b) are Absolute Pose Errors (APEs) for KITTI07 and KITTI10 respectively. (c) and (d) are estimated trajectories on KITTI07 and KITTI10, coded by ATEs.}
	\label{fig:traj_kitti}
\end{figure}


The results are shown in \autoref{tab:eval_kitti}. The translational error comparison demonstrates a comparable localization accuracy of the proposed method against the stereo localizer, while only using temporal monocular information to localize camera in the map. The success tracking on KITTI 01, a highway sequence, indicates the robustness of our system. 
On the other hand, the rotational estimations from our method outperform the stereo localizer. This is because our system better binds the structure from the pre-built map. As we take advantage of temporal multi-view constraints, the variances of localization error are generally much lower than the stereo localizer. 

Additionally, with the comparison of ORB SLAM2 (stereo), improvement of the proposed system to local odometry is noticed. For sequence 04, both localization methods fail to outperform ORB SLAM2 under a structure-less rural environment. This indicates that approaches emphasizing geometry-based matching require diversity of the structure.





\autoref{fig:traj_kitti} illustrates the translation error and trajectory comparison of our approach on KITTI07 and KITTI10. For both sequences, the results exhibit the capability of our method to correct the odometry drift and recover the real scale from the map. For KITTI10 covering a path over 1 km, the localization error is always lower than 1.0 m while the average error is 0.34 m. Large localization error mainly occurrs in structure-less regions, or when the camera is rotated with large angular velocity, for that our algorithms require a structural scene to correct the drift and recover scale for monocular odometry.






\subsection{Local Structure Refinement}
\label{sec:eval_local}

\begin{table}[t!]
	\caption{Comparison of sparse reconstruction RMSE (m) for the proposed method, ORB SLAM (mono) and ORB SLAM (stereo).}
	\begin{center}
		\begin{tabular}{cccc}
			\hline
			     & \textbf{Ours} & \textbf{ORB (mono)} & \textbf{ORB (stereo)}                                                         \Tstrut\Bstrut \\

			\hline

			V101 & 0.0609           & 0.2177     & \textbf{0.0561}    \Tstrut                                                                   \\
			V201 & \textbf{0.0836}           & 0.1176     & 0.1008                                                                      \Bstrut \\
			\hline
		\end{tabular}
		\label{tab:acc_recons}
	\end{center}
\end{table}

\begin{figure}[t!]
	\centering
	\subfloat{\includegraphics[width=0.15\textwidth]{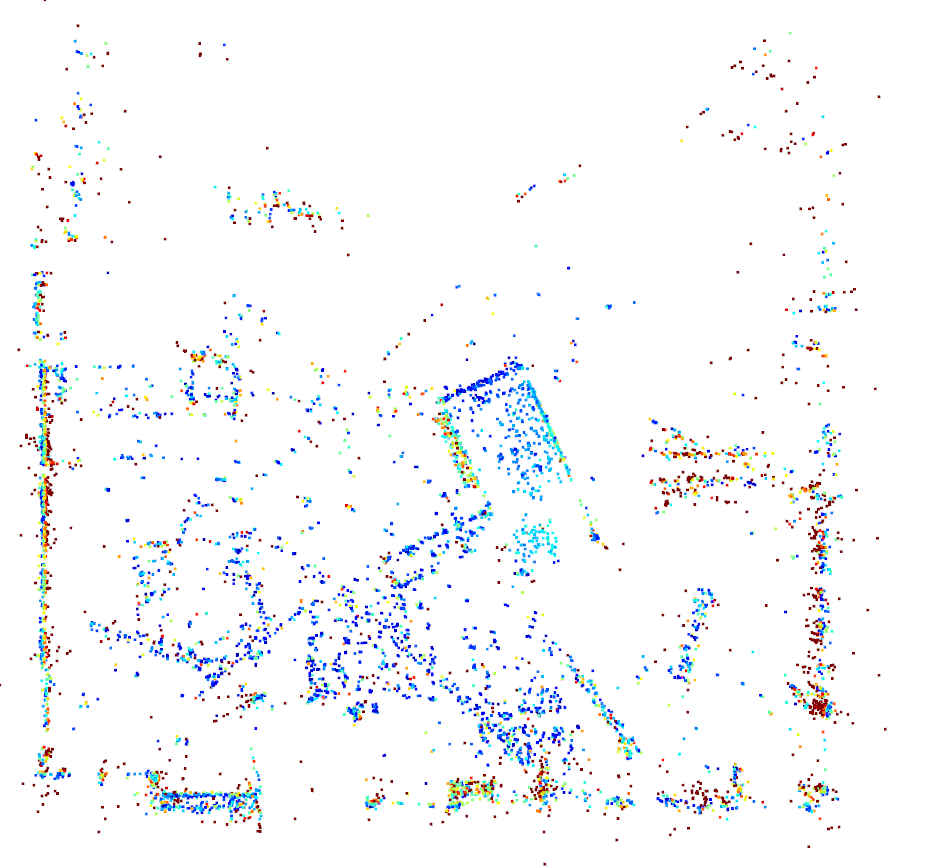}} \subfloat{\includegraphics[height=0.14\textwidth]{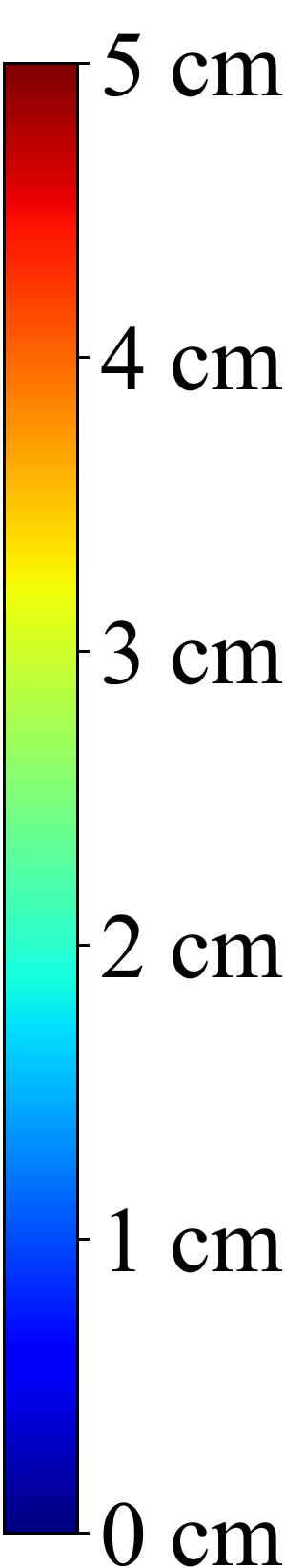}} \qquad
	\subfloat{\includegraphics[width=0.15\textwidth]{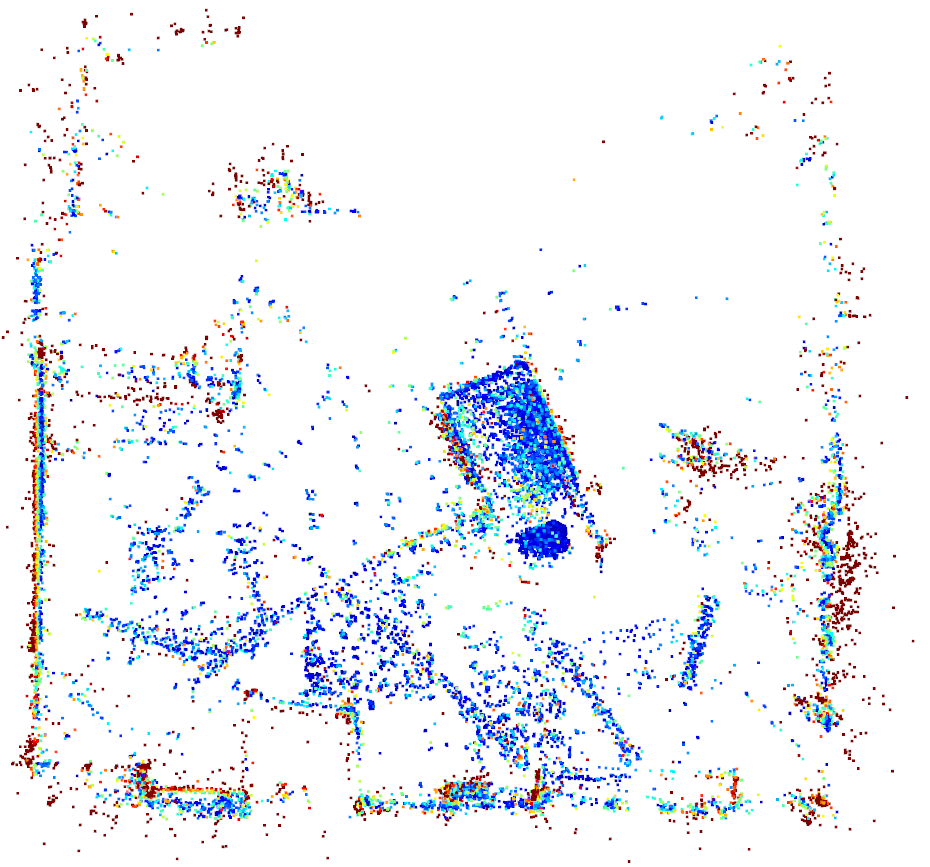}} \subfloat{\includegraphics[height=0.14\textwidth]{fig/recons_colorbar1.pdf}}\\
	\caption{The sparse reconstruction on sequence V201 colored by metric error. The left is from ORB SLAM2 (stereo) and the right is from the proposed method. A comparable local structure to stereo SLAM boosts both tracking and localization, explaining the localization performance of our system. }
	\label{fig:eval_struct}
\end{figure}


To demonstrate the capability of refining local structure, we evaluated the accuracy of the sparse reconstruction on EuRoC Vicon Room sequences. These sequences provide a highly-accurate model from Leica Scan that could serve as ground truth. The sparse feature map reconstructed were saved and then aligned with the global model. For monocular SLAM, the structure was aligned with scale correction. We searched the nearest neighbourhood for each point in the sparse reconstruction and used Euclidean distance between them as the definition of metric error. By calculating RMSE between sparse features and the ground truth, we compare the local reconstruction accuracy against both monocular SLAM and stereo SLAM. The quantitative results are reported in \autoref{tab:acc_recons}. For all the sequences, monocular SLAM system estimates a less accurate structure which in turn leads to less accurate local localization performance. On the contrary, our system achieves the lowest average RMSE below 0.1m, explaining the comparable localization accuracy against the stereo SLAM.

Visualization of structure error is shown in \autoref{fig:eval_struct}. Since our strategy includes features with large distance and narrow baseline across frames, the reconstruction from our system is denser and more outliers are noticed. Despite this fact, our system still achieves a comparable structure accuracy.

\subsection{Evaluation of Initialization Robustness}
\label{sec:eval_init}

\begin{figure}[t!]
	\centering
	\includegraphics[width=0.4\textwidth]{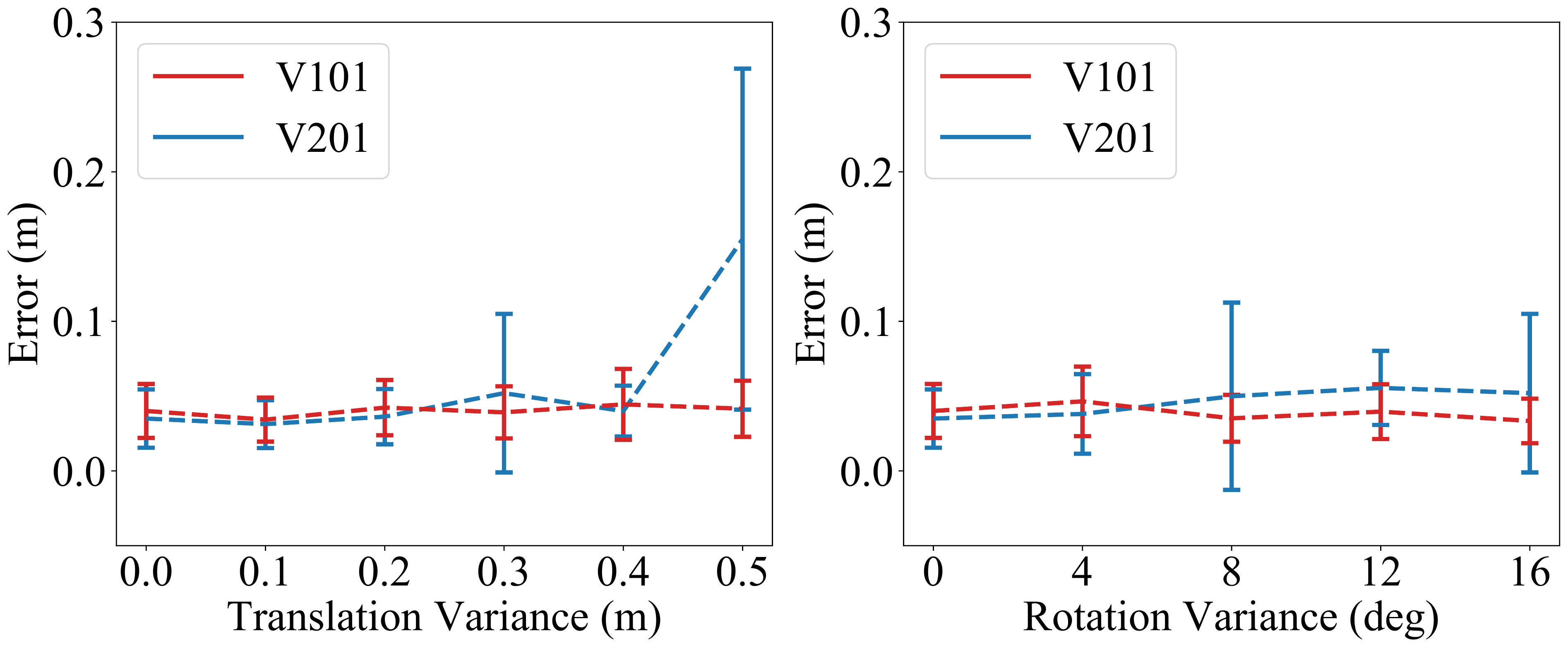}
	\caption{Localization Error with respect to Initialization Variances.}
	\label{fig:err_init}
\end{figure}

\begin{figure}[t!]
	\centering
	\subfloat[\#frames=1]{\includegraphics[width=0.15\textwidth]{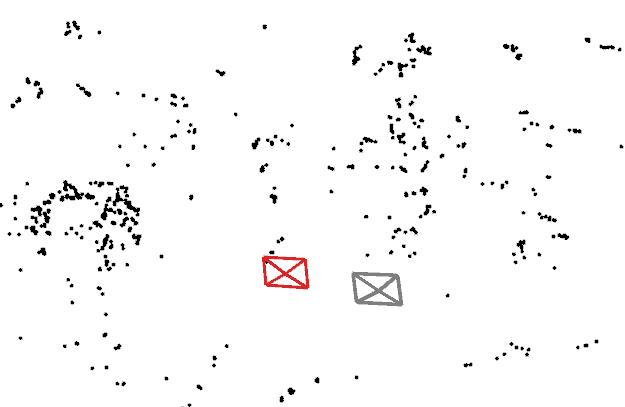}} \enspace
	\subfloat[\#frames=40]{\includegraphics[width=0.15\textwidth]{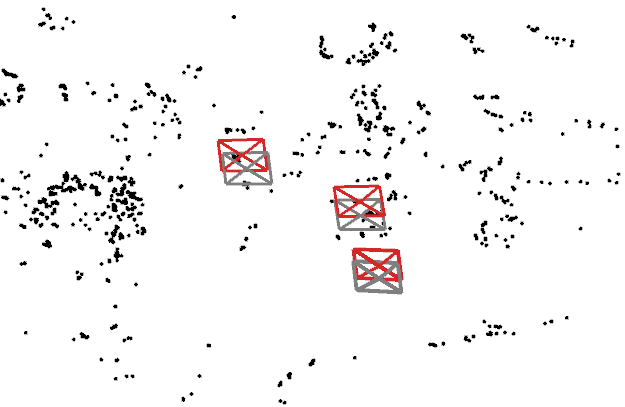}} \enspace
	\subfloat[\#frames=80]{\includegraphics[width=0.15\textwidth]{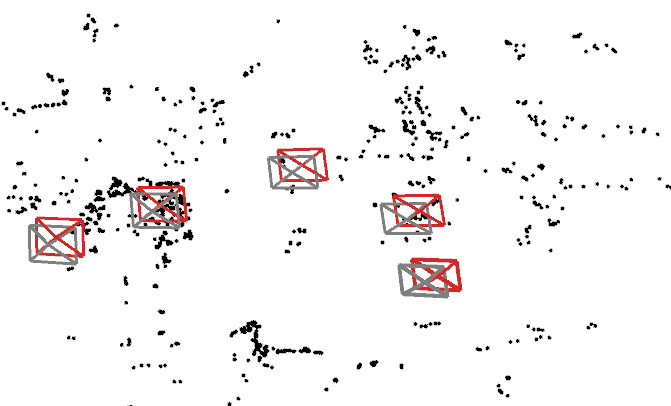}} \enspace
	\caption{A sample initialization process with translational error of 0.25m and rotational error of 10 degree.}
	\label{fig:process_init}
\end{figure}

We evaluated the localization error under different initial poses.
For EuRoC V101 and V201, we added variance to the initial guess along a specific axis for a consistent evaluation.


The quantitative results is shown in \autoref{fig:err_init}, although the initialization variance increases, our system maintains a consistent accuracy under 0.1m with acceptable localization variances, except when translation error is set to 0.5m. As noticed in \autoref{fig:err_init}, localization variance on V201 is larger than on V101. The explanation is that the initial scene of V101 is a degeneration case, a combination of planes, thus optimization direction is typically fixed regardless of initial guess. \autoref{fig:process_init} illustrates how the proposed system recovers from a initialization variance. With one single frame, the structure and absolute scale is ambiguous for the monocular image. 
However, when the scene structure is gradually observed, the local structure is scaled and aligned with the global model, with the depth of features in the first frame being corrected seamlessly. 
Finally, with enough observations, the system converges to a relatively accurate global pose with fewer outliers in the local structure.

\section{Conclusions}
\label{sec:conclusion}


This work presents a hierarchical metric visual localization method based on the SDF-based map representation. 
With a monocular camera, the proposed system exhibits a comparable performance against the state-of-the-art approaches in terms of both accuracy and robustness. 
Further study on the local reconstruction accuracy and robustness on initialization proves the capability of our method to refine local structure for a better global pose estimation.

To achieve a more reliable localization system, we will investigate a more robust initialization strategy such as integrating visual SFM and the SDF observation. 
Another issue of the current SDF-based localization is the large memory usage. We believe a hierarchical map structure is significant for large-scale localization.
Additionally, further study on handling strutural changes might improve the long-term robustness.







\section*{ACKNOWLEDGMENT}
This work was supported by the National Natural Science Foundation of China, under grant No. U1713211, the Research Grant Council of Hong Kong SAR Government, China, under Project No. 11210017, No. 21202816, and the Shenzhen Science, Technology and Innovation Commission (SZSTI) under grant JCYJ20160428154842603, awarded to Prof. Ming Liu.


\balance
\bibliography{ref,datasets,loc,slam,sdf,descriptor}
\bibliographystyle{IEEEtran}

\addtolength{\textheight}{-12cm}   

\end{document}